\documentclass[10pt,twocolumn,letterpaper]{article}
\pdfoutput=1 
\usepackage{cvpr}
\usepackage{times}
\usepackage{epsfig}
\usepackage{graphicx}
\usepackage{amsmath}
\usepackage{amssymb}
\usepackage{algorithm}
\usepackage{algorithmic}
\usepackage{fullwidth}
\usepackage{tabularx}

\usepackage[breaklinks=true,bookmarks=false,colorlinks]{hyperref}

\cvprfinalcopy 

\def\cvprPaperID{5730} 

\ifcvprfinal\fi
\begin{document}

\title{Translate-to-Recognize Networks for RGB-D Scene Recognition}

\author{Dapeng Du \quad Limin Wang\thanks{Corresponding author.}  \quad Huiling Wang \quad Kai Zhao \quad Gangshan Wu\\
State Key Laboratory for Novel Software Technology, Nanjing University, China}

\maketitle
\thispagestyle{empty}

\begin{abstract}
Cross-modal transfer is helpful to enhance modality-specific discriminative power for scene recognition. To this end, this paper presents a unified framework to integrate the tasks of cross-modal translation and modality-specific recognition, termed as Translate-to-Recognize Network {\rm{(\text{TRecgNet})}}. Specifically, both translation and recognition tasks share the same encoder network, which allows to explicitly regularize the training of recognition task with the help of translation, and thus improve its final generalization ability. For translation task, we place a decoder module on top of the encoder network and it is optimized with a new layer-wise semantic loss, while for recognition task, we use a linear classifier based on the feature embedding from encoder and its training is guided by the standard cross-entropy loss. In addition, our TRecgNet allows to exploit large numbers of unlabeled RGB-D data to train the translation task and thus improve the representation power of encoder network. Empirically, we verify that this new semi-supervised setting is able to further enhance the performance of recognition network. We perform experiments on two RGB-D scene recognition benchmarks: NYU Depth v2 and SUN RGB-D, demonstrating that TRecgNet achieves superior performance to the existing state-of-the-art methods, especially for recognition solely based on a single modality.
\end{abstract}


\begin{figure*}
   \centering
      \includegraphics[width=0.98\textwidth]{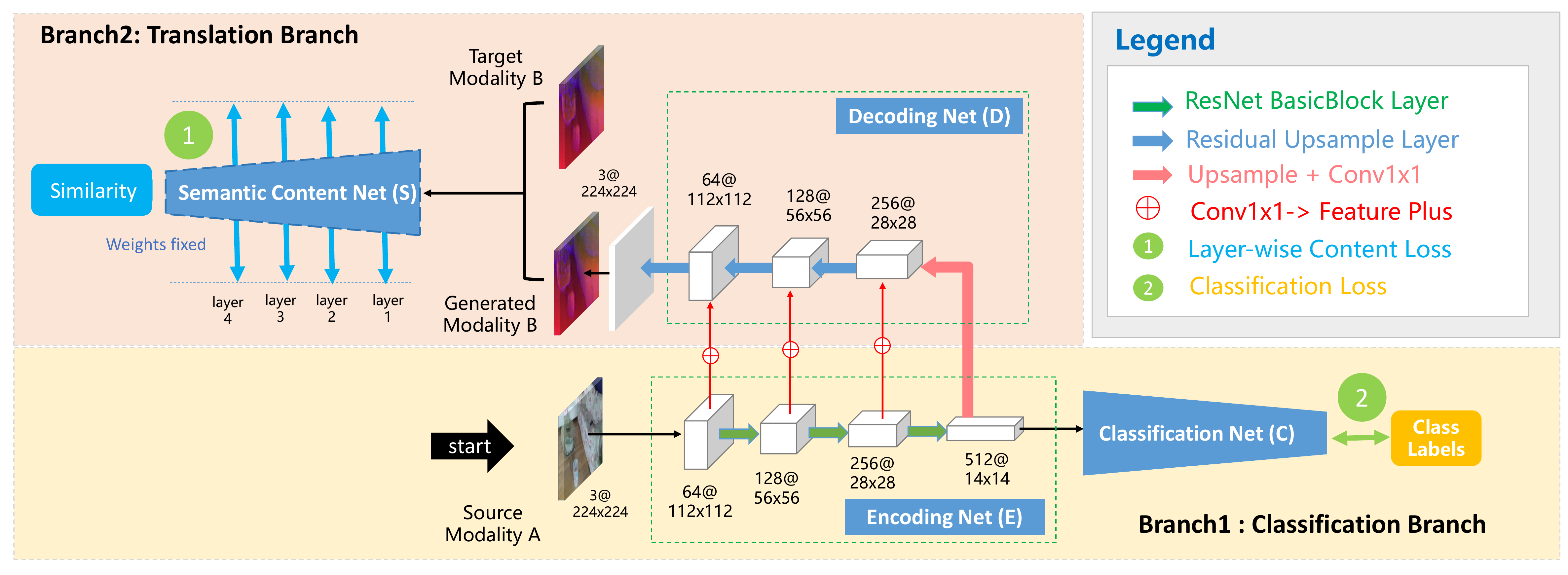}
      \caption{\textbf{TRecgNet.} Cubes are feature maps with dimensions and size represented as \#features@height$*$width. The pipeline consists of two parallel streams: 1) \textbf{Recognition Branch} is for recognizing scene images, in which $E\rightarrow C$ Nets are updated using supervised classification loss. 2) \textbf{Translation Branch} aims at constructing paired complementary-modal data of input through $E\rightarrow D$ Nets. The translation procedure is constrained by semantic supervision $S$ Net. We jointly train the two branches in an end-to-end manner. In the test phase, we only use the Recognition Branch. }
      \label{fig:framework}
\end{figure*}

\section{Introduction}
\label{sec:introduction}

Recently computer vision community has concentrated on applying convolutional neural networks (CNN)~\cite{imagenetkrizhevsky2012} to various vision tasks~\cite{HeZRS16,RenHGS15,HeGDG17,WangGHX017,WangWQG18,WangL0G18}. Meanwhile, the rapid development of cost-affordable depth sensors (\eg, Microsoft Kinect and Intel Realsense) have triggered more attractions to revisit computer vision problems using RGB-D data, such as object detection~\cite{gupta2016cross, xu2017multi}, image segmentation~\cite{mccormac2017scenenet}, activity recognition~\cite{zhang2016rgb, wang2017scene}. 

In this paper, we focus on enhancing the modality-specific network's representative power of RGB-D scene recognition, the goal of which is to accurately classify the given scene images with aligned color and depth information. It is a highly challenging classification task on understanding RGB-D scene data on one hand for its essentially diverse combination of cluttered objects, bounding in different semantic representations for indoors scenes, and importantly on the other hand for the data scarcity problem. The amount of the largest RGB-D dataset~\cite{sunsong2015} is still order-of-magnitude insufficient to provide enough labeled RGB-D data. Since pre-trained RGB CNN models are easily adapted for new RGB data, recent works have focused on learning effective depth features. A few methods~\cite{modalitywang2016, discriminativezhu2016} directly use the pre-trained RGB CNN weights to fine tune a depth CNN, but only limited improvements are given. ~\cite{gupta2016cross} directly transferred semantic supervision from labeled RGB images to unlabeled depth images, which has limits on transform direction. 

This paper tackles major challenges mentioned above from two aspects: 1) We propose to enhance the discriminative power for both RGB and depth's single-modal networks by a cross-modal translation procedure, and 2) we enhance training data sampling with the generated images of high semantic relevance for the classification task. The basic idea is that the modality translation enhances the description power of the encoding network as it forces the RGB/depth data to infer information towards its complementary modality. RGB$\rightarrow$ depth translation network could improve the representation ability of RGB scene network, by learning generating depth data which encodes better on geometric and appearance invariant cues, while depth scene network get benefited learning color and texture cues through depth$\rightarrow$ RGB translation. Meanwhile, this translation process produces new cross-modal data of high quality for the other modality data's classification task. 

Specifically, we propose to couple an arbitrary modality-specific scene~\textbf{Recog}nition network with a modality~\textbf{Tr}anslation network trained in a multi-task manner, termed as~\textbf{TRecg}Net; both branches share the same Encoding Network ($E$ Net), as shown in Figure~\ref{fig:framework}. The effectiveness of TRecgNet that boost modality-specific description power essentially lies on the effect of modality translation, \ie, how the RGB-D data could effectively learn the semantic similarity from the paired data to boost the jointly learning task. We do not simply use the pixel-level Euclidean distance loss as the supervision like many reconstruction works did~\cite{zhan2018unsupervised}. We argue that this difference is unreliable, especially for RGB $\rightarrow$ depth translation, since the ground truth images of depth data usually exist many outliers because of the equipment's limitations and operation errors. Also, low level pixel-wise similarity fails to provide any semantic relevance. Instead, we perform the translation process using pre-trained semantic model, which is inspired by the style transfer related works~\cite{gatys2016image, johnson2016perceptual, chen2018cartoongan}. It has been shown that CNNs trained with sufficient labeled data on specific vision tasks, such as object classification, has already learned to extract semantic content representations. This generalized ability is not limited to specific datasets or tasks. In these works, authors use the perceptual constraint from one specific layer, usually conv4\_x of VGG model, to keep the "rough" content of the image when transforming it to another style. In contrast, we propose to leverage perceptual loss from multiple layers supervising the translation process, for a simple yet effective intuition that higher layer tends to preserve the semantic content while lower one does better in capturing the detailed information, which could provide enough and effective cues for cross-modal similarity learning.

We test the TRecgNet on the two benchmarks of the RGB-D Indoor scene recognition task, SUN RGB-D dataset~and NYU Depth dataset v2. Our TRecgNet could obtain an evident improvement on both modality-specific and RGB-D settings. The main contributions of this paper are two-fold: 

\begin{itemize}
\item A TRecgNet is proposed to transfer complementary cues through a label-free modality translation process, which improves modality-specific classification task in an end-to-end manner and achieves state-of-the-art performance on RGB-D indoor scene benchmarks. 
\item TRecgNet can generate more photo-realistic and semantically related data to enhance training data sampling which alleviates the data scarcity problem and effectively improves the classification performance.
\end{itemize}

\section{Related Work}
\textbf{RGB-D Scene Recognition.}
Earlier works relied on handcrafted features to capture the characteristic properties of the scene. Banica~\etal~\cite{banica2015second} used second-order pooling to quantize local features for segmentation and scene classification. Gupta~\etal~\cite{gupta2015indoor} proposed to quantize segmentation outputs by detecting contours from depth images as local features for scene classification. Recently, multi-layered networks such as CNNs is able to learn useful representations from large amounts of data. In general, they learned modality-specific features from RGB and depth images separately, and then performed the fusion. Wang~\etal~\cite{modalitywang2016} extracted and combine deep discriminative features from different modalities in a component aware fusion manner. Gupta~\etal~\cite{gupta2016cross} transferred the RGB model to depth net using unlabeled paired data according to their mid-level representations. However, these methods only consider transferring color cues to depth net and ignore that depth cues could also benefit the RGB net. More recently, the recognition power of depth net has been comprehensively studied. In~\cite{song2018learning}, Song~\etal argued that learning depth features from pre-trained RGB CNN models are biased and seek to learn depth features from scratch using weakly supervised depth patches. However, since depth data still suffers from data scarcity problem, the structure of the depth nets fail to go deep which limits its extendibility .

\textbf{Paired Image-to-Image Translation.}
Paired image-to-image translation problems could be formulated as pixel level mapping functions, however, each pixel is translated independently from others in this manner~\cite{zhang2016colorful}. Recently, GANs~\cite{goodfellow2014generative,unsupervisedradford2015} have achieved impressive results in image generation. Wang~\etal~\cite{generativewang2016} factored the generation into two steps: surface normal generation and content generation. They used a pixel-wise surface normal constraints as additional supervision. In [18], the authors proposed to use GANs to learn the mapping functions between paired images. These GAN constraints help generate more variegated images from data distribution learning, however, GAN based model is hard to train and generated images tend to lack the semantic consistency of source images that are hard to be leveraged in subsequent semantics related work, such as classification or segmentation tasks. More recently, in many style transfer works~\cite{gatys2016image, johnson2016perceptual, chen2018cartoongan}, they use perceptual loss from specific layer of pre-trained RGB models, to maintain the structural content during the translation. However, most of them only use content constraint from one specific layer, usually conv4\_x of VGG model, ignoring that pre-trained models could perform different levels of semantic supervisions which are good enough for image translation. In this paper, we propose to leverage perceptual loss from multiple layers to constrain the translation process of RGB-D data. Unlike pure image generation task, we focus on how this translation process finally benefits the classification task and the performance of data augmentation using generated images.  

\section{Method}
This section details the description of our Translate-to-Recognize Network (TRecgNet). The proposed framework is illustrated in Figure~\ref{fig:framework}. Assume that in RGB-D setting we would like to train a classification network using one modality data. Let $(\mathbf{M_{A}}, \mathbf{M_{B}})$ be the paired RGB-D images of specific classes from set $L=\{1, 2, ..., N_{c}\}$, where $N_{c}$ is the total number of scene classes. Our object is to learn an embedding $E:\mathbf{M_{A}}\rightarrow \mathbb{R}^{d}$ with a translation mapping $T:\mathbb{R}^{d}\rightarrow \mathbf{M_{B}}$ and a class prediction function $ C:\mathbb{R}^{d}\rightarrow \mathbf{L}$. The core problem is how the translation process could make the modality-specific Encoder Network ($E$ Net) learn effective complementary-modal cues to benefit the classification task. 
\begin{figure}
	\includegraphics[width=0.47\textwidth]{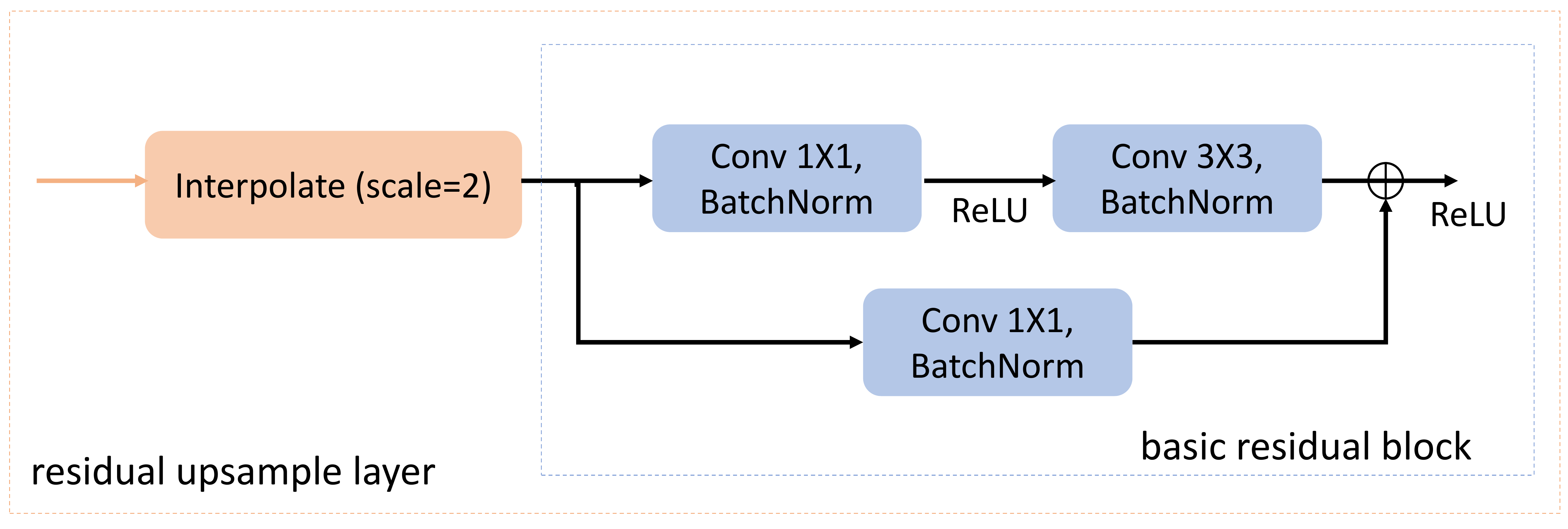}
	\caption{Architecture of the residual upsample layer. The input feature maps get upsampled with scale 2 followed by one basic residual block.}
	\label{fig:residual-upsample}
\end{figure}
\subsection{The TRecgNet Architecture}
\label{sec:arch}

TRecgNet consists of four parts including Encoder Network~($E$ Net), Classification Net~($C$ Net), Decoder Net~($D$ Net) and Semantic Content Network~($S$ Net). Figure~\ref{fig:framework} shows the exemplar architecture of RGB TRecgNet built on ResNet-18, which uses pre-trained ResNet18 as the $E$ and $C$ Nets. $D$ decodes the feature maps from $E$ to reconstruct the complementary data. Depth TRecgNet uses the same structure only by exchanging the position of source and target modalities. We make three measures to enhance the translation process. First, we empirically remove the first max-pooling operation of ResNet. In the whole structure, feature maps only shrink by convolution operation with stride and less information would get lost which is very important for image translation process. Second, we introduce residual upsample layer in the $D$ Net. A residual upsample layer upsamples feature maps with a bilinear interpolation operation accompanied with one residual block which mimics the basic residual block of ResNet. Figure~\ref{fig:residual-upsample} shows the architecture of the residual upsample layer. Third, we propagate the context information from three stages of $E$ to the corresponding outputs of $D$ similar with~\cite{ronneberger2015u}. We use the plus operation instead of concatenation which could reduce the number of parameters in the $D$ Net. 
\subsection{Layer-wise Content Supervised Translation}
\label{sec:translation}

The translation network aims to improve the presentation ability of $E$ Net learning characteristics of complementary data, the procedure of which is supervised by a semantic RGB CNN model, which uses the combination of perceptual constrains from lower layers to higher ones, measuring the \textbf{layer-wise similarity} of the generated and the paired data. 

We use the ResNet model pre-trained on ImageNet~\cite{zhou2017places} as the supervision content network $S$, for the consideration of accordance with the architecture of $E$ Net. More details could be referred to Section~\ref{sec:arch}. We denote the image representations of $S$ Net as $\Phi=\{\phi_{M}^l, l\in[1,2,3,4] \}$, where $\phi_{M}^i$ is the $i^{th}$ layer representation for input data M. It maps an input image from modality $M$ to a feature vector in $\mathbb{R}^{d}$. Specifically, we define the $L1$ loss between two feature vectors for translation supervision. Suppose we are training the classification task for $\mathbf{M_{A}}$. Generated images ${y_i'}$ and $\mathbf{M_{B}}$ are fed into the $S$ Net, and we can get layer-wise presentations from ${y_i'}$ and $\mathbf{M_{B}}$. We constrain them from every each layer (layer1-layer4 in ResNet) be the same by $L1$ loss:

\begin{eqnarray}
\label{eqn:lossContent}
L_{content}(y_i, y_i', l)=\sum_{l=1}^L\Vert\phi_{y_i}^l-\phi_{T(x_i)}^l\Vert_{1} 
\end{eqnarray}

\subsection{Training Strategy}
In this section, we introduce our optimization procedure in detail. To jointly learn the embedding and the translation pair, we optimize $E, C, D$ networks in a multi-task manner. Specifically, given a pair of RGB-D images, let $e_{\mathbf{M_{A}}} = E(\{x_i\})$ be the embeddings from $E$ Net computed on $\mathbf{M_{A}}$ and $d_{\mathbf{M_{B}}} = D(E(\{x_i\})$ be the generated modality B data decoded from $D$. We simultaneously update 1) the $E\rightarrow D\rightarrow S$ to minimize the distance of layer-wise vectors to constrain their semantic similarity, and 2) the $E\rightarrow C$ using a cross entropy loss function for classification task. The total loss is updated with a linear combination: 

\begin{align}
& L_{total}=\alpha L_{content}+\beta L_{C_{cls}}
\end{align}

\noindent where $L_{content}$ is Equation~(\ref{eqn:lossContent}) described in Section~\ref{sec:translation}, $L_{cls}$ is the cross-entropy loss for classification task, with their coefficients $\alpha, \beta$ set as 10, 1 from the best trials. 

As the datasets we use in Section~\ref{sec:exp} are characterized by unbalanced images for categories, we use rescaling weights given to each category for cross-entropy lossRescaling weights aims to assign different weights to different classes to handle the issue of imbalance training. Specifically, we use the following rescaling strategy:

\begin{eqnarray}
L_{weighted\_cls}  = \frac{1}{N}\sum_i-w(y_i)\log\frac{f(x_i)_{y_i}}{\sum_j f(x_i)_j}.
\end{eqnarray}
The weight $w(y)$ is computed as:
\begin{eqnarray}
w(y)=\frac{N(y)-N_(c\_min)+\delta}{N_(c\_max)-N_(c\_min)},
\end{eqnarray}
where $N_{y}$ is the number of images of class $y$. $c\_min$ and $c\_max$ represent the class with the least and the most number of training images. The $\delta$ is set as 0.01. 

In the test phase, we only use the recognition branch for recognition prediction, as illustrated in Figure~\ref{fig:framework}. 

\textbf{Imbalanced translations between two modalities.}
There are several factors demonstrating that the translations from RGB-to-depth and depth-to-RGB are imbalanced. For example, translation from RGB to depth images is more relatively natural procedure while it would become an ill-posed problem from the reversed direction. What's more, the ground truth of depth data comes with much more value errors due to the characteristics of collecting equipments and process. Therefore, we sample a random noise vector from N(0,1) concatenated to the input feature to $D$ in the Depth TRecgNet (translation from depth to RGB). We found it useful to stable the training of Depth TRecgNet. The generated RGB images also get an interpretable control from the sampled noise. The dimensionality of noise is set as 128 in our experiments.
\begin{figure}
	\centering
	\includegraphics[width=0.47\textwidth]{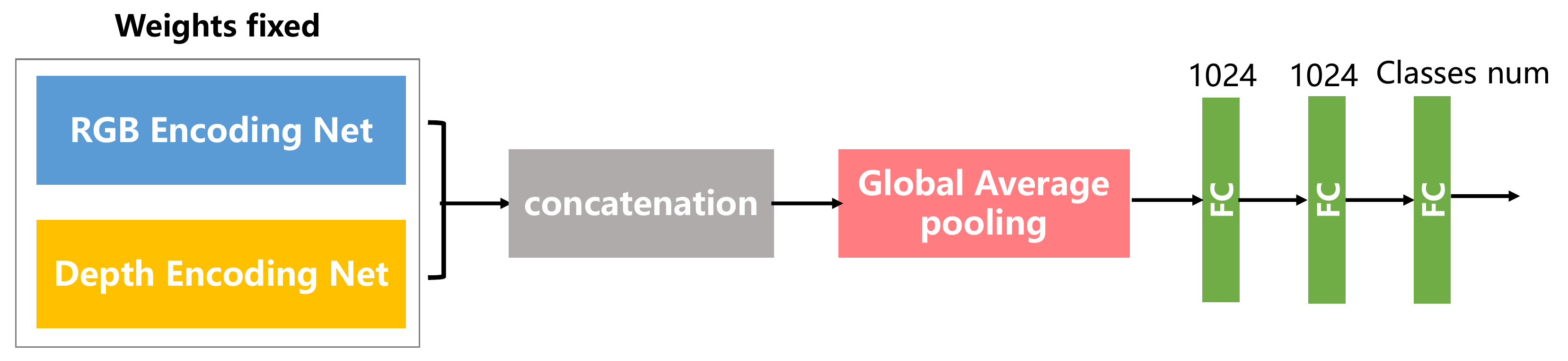}
	\caption{Fusion Network. We only use Encoding Nets from RGB and Depth TRecgNets for fusion. The weights of two $E$ Nets are fixed, and only weights of the classifier are updated.}
	\label{fig:fusion}
\end{figure}

\textbf{Initialization with unlabeled RGB-D data.}
As mentioned in Section~\ref{sec:introduction}, the sizes of most existing labeled RGB-D datasets are in small orders of magnitude compared with RGB datasets. However, there are a large number of unlabeled RGB-D pairs, for example, from RGB-D video sequences. A significant advantage of our method is that we are capable of initializing the TRecgNet with these unlabeled data. In other words, the modality translation process is a label-free procedure, by which the TRecgNet could learn rich representations from unlabeled RGB-D data boosting the further task. Related experiments are detailed in Section.~\ref{sec:exp}. 
\subsection{Fusion}
After we get two trained TRecgNets for RGB and depth data, we compute two $E$ Nets and concatenate modality-specific features from them. The embedding is operated on global average pooling (GAP)~\cite{lin2013network} to reduce the number of parameters followed by three fully connected layers. The whole structure is illustrated in Figure~\ref{fig:fusion}. We fix the Encoding Networks and directly train the classifier in an end-to-end manner. We find this would be superior to that directly combines the two prediction results and show the effectiveness of modality-specific networks more straightforwardly.
\section{Experiments}
\label{sec:exp}
In this section, we first introduce the evaluation datasets and the implementation details of our proposed approach. Then we discuss the ablation study of TRecgNet on effectiveness and layer contribution of $S$. We also compare the quality of generated images with other approaches. Finally, we evaluate the performance of our approach with state-of-the-art methods. We quantitatively report the average accuracy over all scene categories following the conventional evaluation scheme.
\subsection{Datasets}
\textbf{SUN RGB-D Dataset} is currently the largest RGB-D dataset. It contains RGB-D images from NYU depth v2, Berkeley B3DO~\cite{categoryjanoch2013}, and SUN3D~\cite{sun3dxiao2013} and is compromised of 3,784 Microsoft Kinect v2 images, 3,389 Asus Xtion images, 2,003 Microsoft Kinect v1 images and 1,159 Intel RealSense images.  Following the standard experimental settings stated in~\cite{sunsong2015}, we only work with 19 major scene categories which contain more than 80 images. As per standard splits, there are in total 4,845 images for training and 4,659 for testing.

\textbf{NYU Depth Dataset V2 (NYUD2)} is a relatively small dataset; only a few of its 27 indoor categories are well presented. Following the standard split in~\cite{indoorsilberman2012}, the categories are grouped into ten including nine most common categories and an \textit{other} category representing the rest. Also, we use 795 / 654 images for training / testing following the standard split.
\begin{table}
	\begin{center}
	\begin{tabularx}{.48\textwidth}{|X|c|c|c|}
		\hline
			Data  & Model & Init &Acc (\%)\\
		\hline 
			          & ResNet18 & Places &47.4\\	
		
			  RGB    & TRecg & Places &49.8\\
		
			          & TRecg Aug & Places &50.6\\
		\hline
			          & ResNet18 & random &38.1\\
		
			          & ResNet18 & Places &44.5\\

			          & TRecg & random &42.2\\

			 Depth   & TRecg & Places &46.8\\

			          & TRecg & random/unlabeled &44.2\\
	
			          & TRecg & Places/unlabeled &47.6\\

			          & TRecg Aug  & Places  &47.9\\
		\hline
	\end{tabularx}
	\end{center}

	\caption{\textbf{Ablation study of TRecg-ResNet18 on recognition performance}. The results are reported on the test set of SUN RGB-D (Top-1's mean accuracy \%). ``Aug'' means using generated data in training.}
	\label{tab:ablation}
	
\end{table}
\subsection{Implementation Details}
The proposed approach is implemented in popular deep learning framework, Pytorch~\cite{paszke2017automatic}, on an NVIDIA TITAN Xp GPU. We train the network using Adam stochastic optimization~\cite{kingma2014adam} to learn network parameters, with the batch size set to 40. The RGB-D images are resized to $256\times256$ and randomly cropped to $224\times224$. We train the TRecgNet in 70 epochs, and the learning rate is initialized as $2\times10^{-4}$ at the first 20 epochs and linearly decades in the rest of 50. In the test phase, we use a center crop operation on test images. We employ geocentric HHA (Horizontal disparity, Height above ground and Angle with gravity)~\cite{gupta2014learning} to encode depth images, which has been shown better performance to capture the scenes structural and geometrical properties of depth data for kinds of vision tasks. In the subsequent experiments, we separately train two kinds of TRecgNets for evaluation. The basic TRecgNet is trained without using generated data while the TrecgNet Aug refers to training a TRecgNet with generated data from the previous corresponding basic TRecgNet. (RGB TRecgNet Aug leverages the basic depth TRecgNet as the generated data sampler, and vice versa). Specifically, in the training phase of TrecgNet Aug, we randomly use the generated data, the number of which is controlled as 30\% of batch size, to achieve the best performance. 

\begin{figure}
	\centering
	\includegraphics[width=0.47\textwidth]{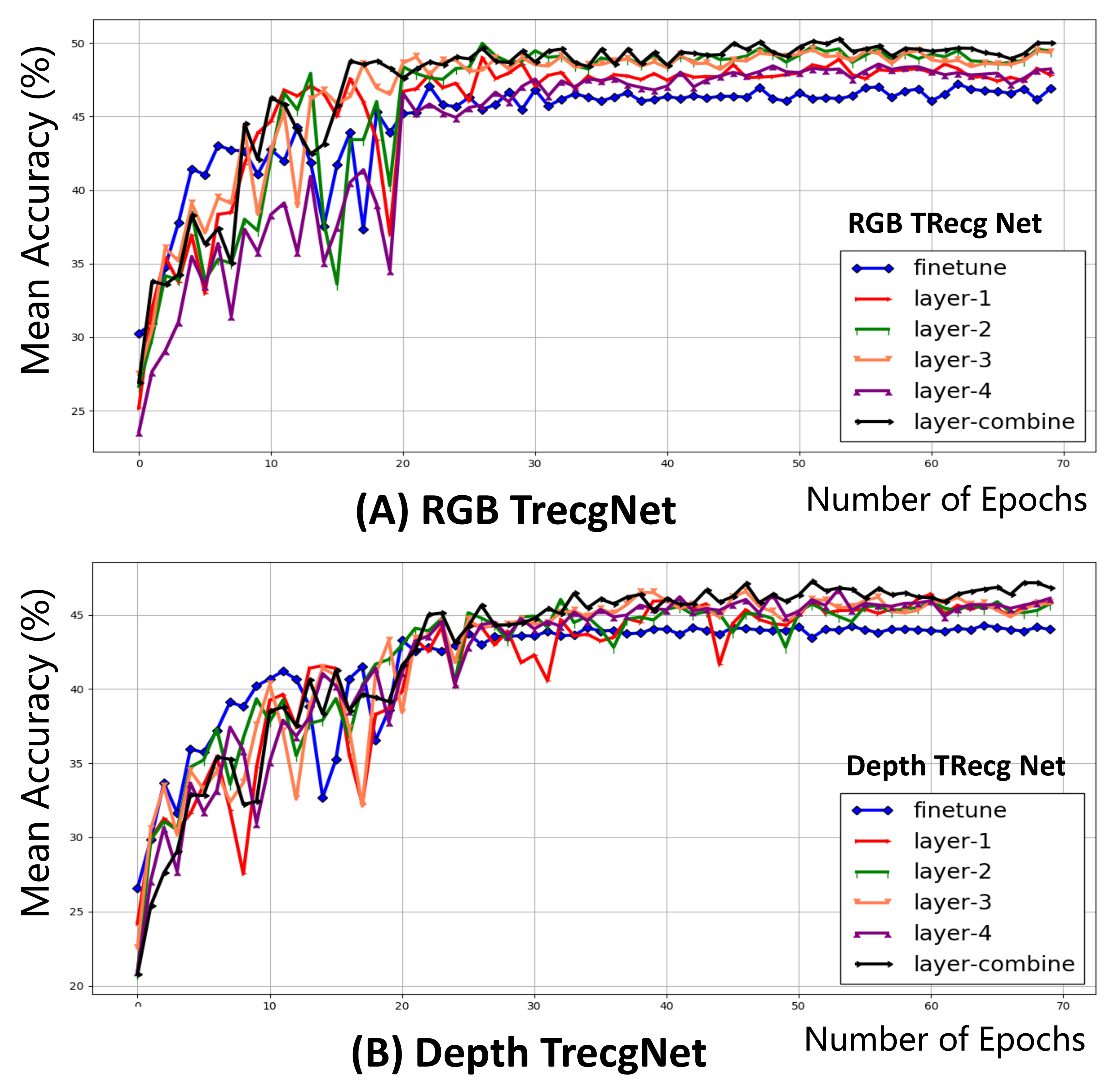}
	\caption{Effect of content model using different layer as supervision on classification for (A)~\textbf{RGB} TRecgNet and (B)~\textbf{Depth} TRecgNet. Tested on SUN RGB-D dataset.}
	\label{fig:layer-acc}
\end{figure}

\begin{figure}
	\includegraphics[width=0.47\textwidth]{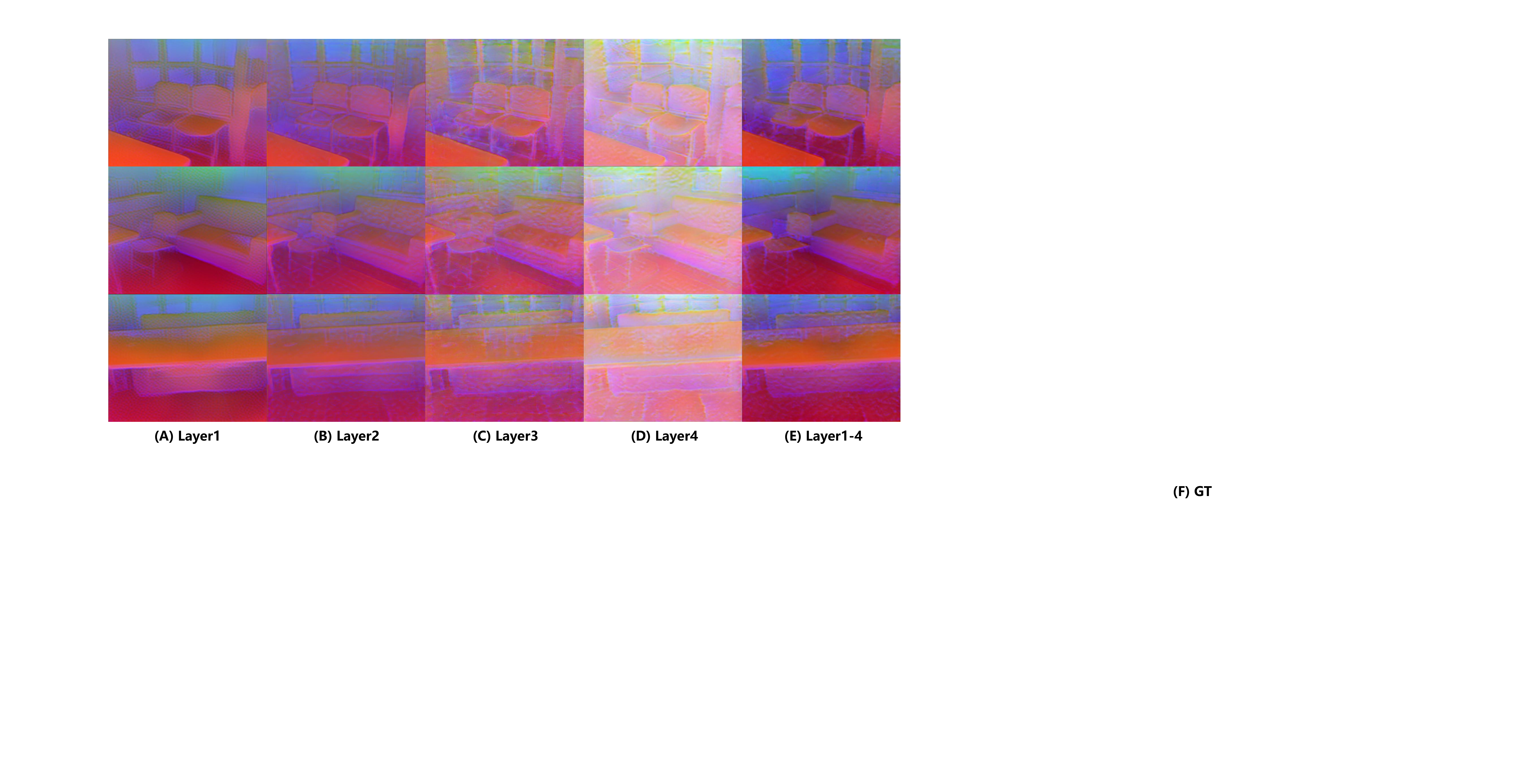}
	\caption{Examples of TRecgNet translating RGB images to depth ones by semantic supervision from different layers. The combination of layer-wise content supervision gives the best photo-realist translation. The input images are all from the test set of SUN RGB-D dataset.}
	\label{fig:layer-wise-display}
\end{figure}
\begin{figure*}
	\centering
	\includegraphics[width=0.95\textwidth]{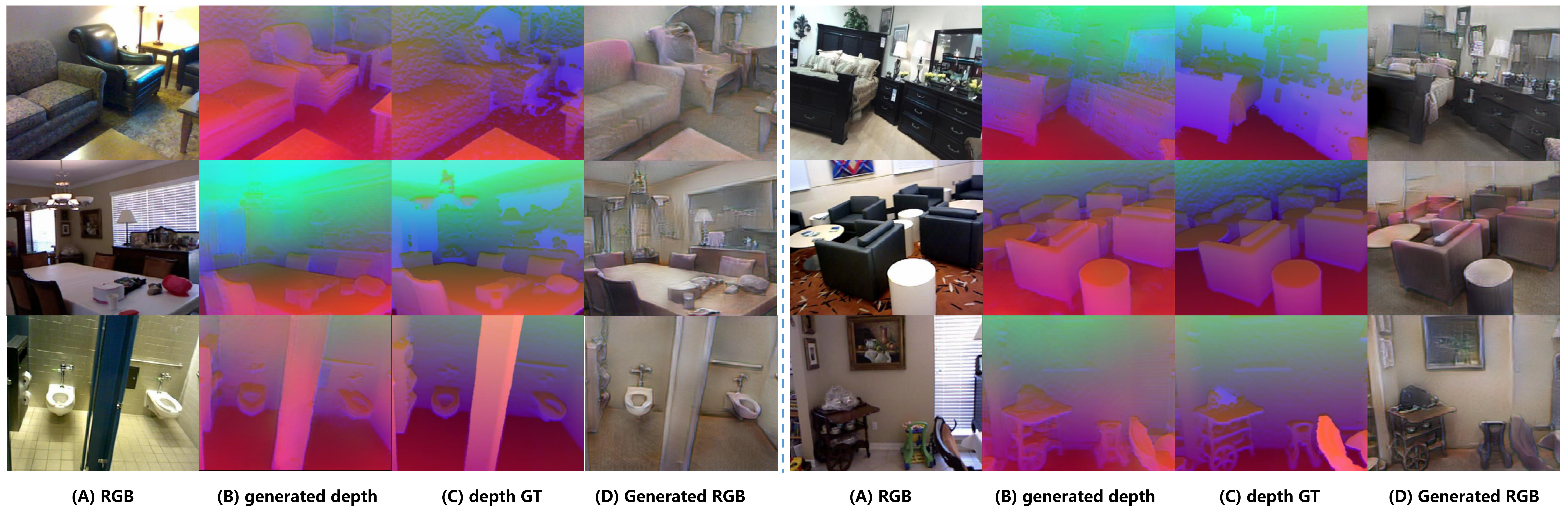}
	\caption{Examples of generated data by our TrecgNet from test set of SUN RGB-D dataset. (B) are depth images translated from original RGB data (A), and (D) are generated RGB images using original depth ones (C). }
	\label{fig:generated-images}
\end{figure*}
\subsection{Results on SUN RGB-D dataset}
\textbf{Study on the effectiveness of TRecgNet.} We begin our experiments by studying the effectiveness of TRecgNet for RGB-D recognition task. We tend to prove that 1) learning essential similarity from modality translation branch could effectively benefit classification task, and 2) Unlabeled RGB-D data helps in training TRecgNet for recognition task, and 3) generated data improves training process. Since patterns of color and depth information vary greatly on visual appearance, geometry and surface, we use the following baselines: for the classification of RGB modality, we fine tune pre-trained ResNet18 on RGB images; for the depth modality, we train the network both from scratch and from the pre-trained model.


\begin{table}
	\begin{center}
	\begin{tabularx}{.48\textwidth}{|X|c|c|c|}
		\hline
			Loss         & Pixel& Pixel+GAN& Ours\\
		\hline 
			Mean Acc (depth)     & 13.1\%     & 17.7\%         & 30.3\%\\
		\hline 
			Mean Acc (rgb)     & 7.3\%     & 20.6\%         & 18.4\%\\
		\hline
	\end{tabularx}
	\end{center}

	\caption{We compare the quality of generated depth images by training a vanilla ResNet18 pre-trained on Places dataset. The training images are all generated using training data of SUN RGB-D dataset by three methods. The results are reported on the test set of SUN RGB-D (Top-1's mean accuracy \%). Images generated by our TRecgNet achieve the best result.}
	\label{tab:gen_acc}
	
\end{table}

We test our TRceg Net with following settings: a) the same initialization schedule with baseline, b) the Depth TRecgNet is pre-trained using 5k unlabeled RGB-D video sequences from the NYUD2 dataset. c) Two TRecgNets extract the translation branches as the data augmentation supplier to retrain the TRecgNet;~30\% of original training data is randomly replaced with generated data. The experimental results are summarized in Table~\ref{tab:ablation}. We observe that for each modality, our TRecgNet outperforms the baseline with big margins. RGB TRecgNet outperforms the baseline by 2.4\% while Depth TRecgNet by 3.7\% and 2.3\% with randomly initialization and pre-trained weights from Places dataset, separately. 

It is worth noting that unlabeled RGB-D data pre-training for TRecgNet makes a further boosting of 2\% and 0.8\%, when training from scratch and using pre-trained weights, respectively. Without pre-trained Places weights but only 5k unlabeled data, the result is very comparable to that of vanilla ResNet pre-trained on Places dataset could achieve, only with quite less training data. This indicates that we could flexibly design a $E$ Net achieving acceptable results without pre-training it using large-scale datasets like ImageNet or Places dataset. 

 We also find promotions when using generated images as training data, with 1.1\% for the Aug-Depth and 0.8\% for Aug-RGB TRecgNets. We show some generated data examples in Figure~\ref{fig:generated-images}. It's interesting to find that the generated depth images tend to be brighter and have better contours than the original ones, due to the learned RGB context information in the translation process. As for generated RGB data, since the original depth data exists with non-negligible measurement error, generated RGB data would be inevitably of low quality on details, however, we turned out to find it be alleviated by adding the random noise in the training of Depth TRecgNet. 
\begin{table*}
	\begin{center}
	\begin{tabularx}{.95\textwidth}{p{1.5cm}Xccp{1cm}p{0.8cm}p{0.8cm}}
		\hline\hline
			 &  &  &   & \multicolumn{2}{c}{Accuracy (\%)}  &    \\
			& Method & RGB Init & Depth Init  & RGB & Depth& Fusion\\
        \hline
		
		    Baseline & ResNet18 & ImageNet & ImageNet & 46.6 & 44.5 & 50.1\\
		             & ResNet18 & Places & Places & 47.4 & 44.8 & 50.8\\

		\hline
		    		 & TRecgNet & ImageNet & ImageNet & 48.7 & 46.9 & 55.5\\
		 	Proposed & TRecgNet Aug & ImageNet & ImageNet & 49.2 & 47.9 & \textbf{56.1}\\
		 			 & TRecgNet & Places & Places & 49.8 & 46.8 & 56.1\\ 
		 			 & TRecgNet Aug & Places & Places & 50.6 & 47.9 & \textbf{56.7}\\
		\hline
					
				     & Multimodal fusion\cite{discriminativezhu2016} & Places & Places & 40.4 & 36.5 & 41.5\\
					    
SOTA & Modality\&component fusion~\cite{modalitywang2016} & Places & Places & 40.4 & 36.5 & 48.1\\

					 	 & RGB-D-CNN+wSVM~\cite{song2018learning} & Places & Fast R-CNN & 44.6 & 42.7 & 53.8\\
					 	 & DF$^{2}$Net metric learning~\cite{Li2018DF2NetDF} & Places & Places & 46.3 & 39.2 & 54.6\\
					 			 
     \hline
					 
	\end{tabularx}
	\end{center}
	\caption{Comparison with state-of-the-art methods on the test set of the \textbf{SUN RGB-D} dataset. The performance is measured by Top-1's mean accuracy over classes. ``Aug'' means using generated data in training.}
	\label{tab:sun-rgbd}

\end{table*}
\begin{table*}
	\begin{center}
	\begin{tabularx}{.95\textwidth}{p{2.5cm}Xccp{0.8cm}p{0.8cm}p{0.8cm}}
		\hline\hline
			 &  &  &   &  \multicolumn{2}{c}{Accuracy (\%)} &    \\
			 & Method & RGB Init & Depth Init  & RGB & Depth& Fusion\\
      \hline
		   Baseline & ResNet18 & Places & Places & 59.8 & 52.3 & 63.8\\
						
		\hline
				 & TRecgNet & Places & Places & 60.2 & 55.2 & 65.5\\
		Proposed & TRecgNet & SUN RGB-D & SUN RGB-D & 63.8 & 56.7 & 66.5\\
				 & TRecgNet Aug & SUN RGB-D & SUN RGB-D & 64.8 & 57.7 & \textbf{69.2}\\
					 
		\hline
									  					    
 				& Modality \& component fusion~\cite{modalitywang2016} & Places & Places &  53.5 & 51.5 & 63.9\\

SOTA & RGB-D-CNN+wSVM~\cite{song2018learning} & Places & Fast R-CNN & 53.4 & 56.4 & 67.5\\

					 	 & DF$^{2}$Net metric learning~\cite{Li2018DF2NetDF} & Places & Places & 61.1 & 54.8 & 65.4\\
					 			 
     \hline
					 
	\end{tabularx}
	\end{center}
	\caption{Comparison with state-of-the-art methods on the test set of the \textbf{NYUD2} dataset. The performance is measured by Top-1's mean accuracy over classes. ``Aug'' means using generated data in training.}
	\label{tab:nyud2}

\end{table*}

\textbf{Study on layer contribution of Semantic Content Network.}
The effectiveness of transferring complementary cues relies much on the semantic content model $S$ Net. Therefore, we take an interest in how the $S$ Net affects the translation and recognition tasks using the different layer as supervision. We separately use layer 1 $\sim$ layer 4 as well as their combination as the translation constraint to test the scene recognition performance of TRecgNet, compared with directly fine tuning pre-trained ResNet18 model; all the experiments are conducted on SUN RGB-D dataset. In Figure~\ref{fig:layer-acc}, the accuracy is plotted against the number of training epochs. The following observations can be made: 1) TRecgNet achieves markable improvement to varying degree by using the different layer as supervision. 2) The training procedure becomes more steady when using layer-wise content constraint and achieves the best performance. The result demonstrates the effectiveness of TRecgNet using $S$ as the supervision for cross-modal translation. Figure~\ref{fig:layer-wise-display} gives an example of RGB TrecgNet translating images by semantic supervisions from different layers. Similarity learning from lower layers' supervision focus on keeping textures while higher layers' constraints try to capture more semantic cues ignoring luminance or texture. In contrast, layer-wise content supervision combines multi-layer characteristics which could not only get better performance on classification task but generate more photo-realistic images.

\begin{figure*}
	\centering
	\includegraphics[width=0.82\textwidth]{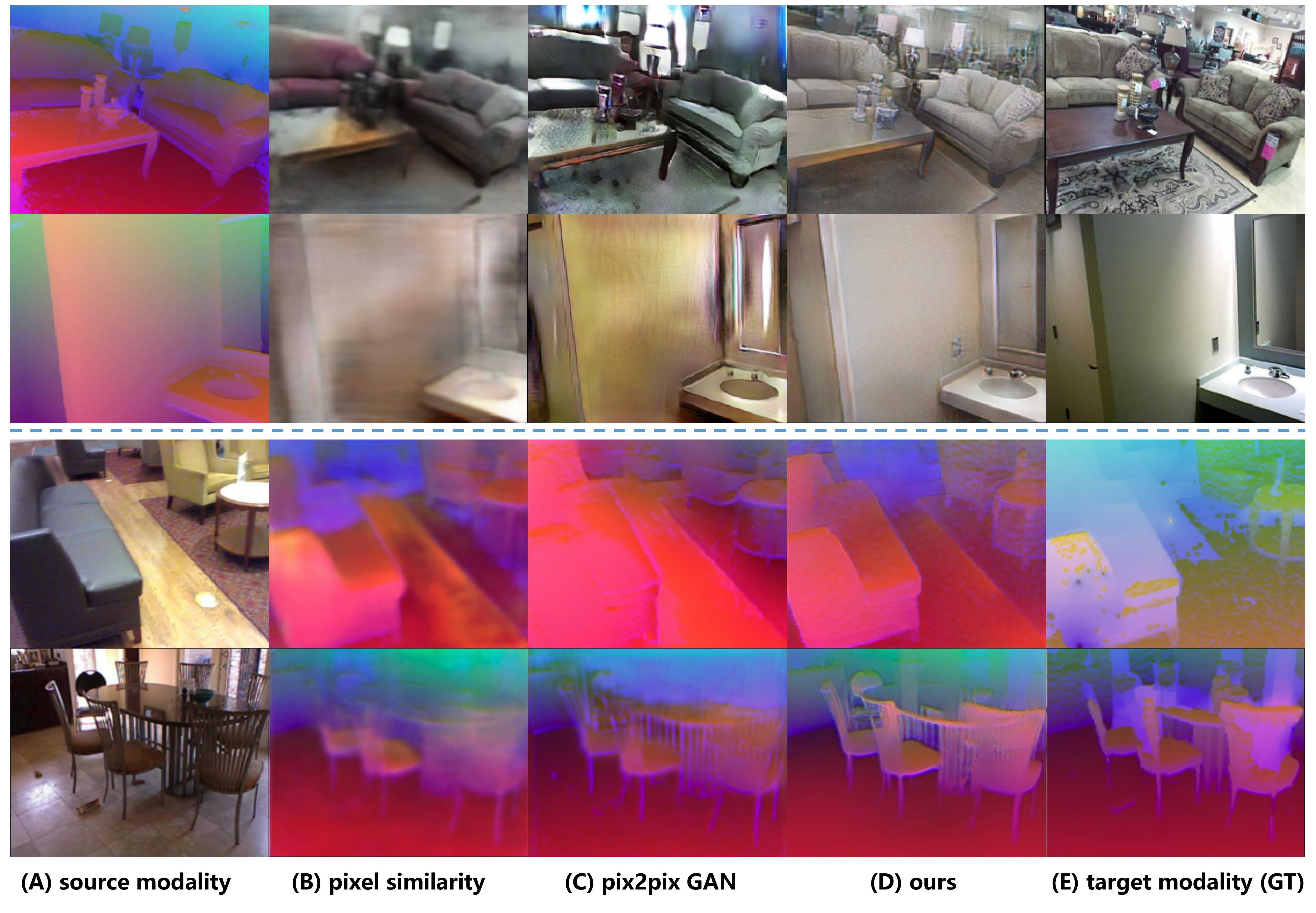}
	\caption{Generated examples by different methods. RGB data translated to depth one is shown in the first two rows while images in the last two rows give a reversed example. (A), (E) are ground truth of RGB and depth data. (B), (C) and (D) are generated images separately from pixel-level L1 loss supervision, pixel-to-pixel GAN, and our TRecgNet.}
	\label{fig:compare-images}
	\vspace{-2mm}
\end{figure*}

\textbf{Study on quality of generated images.} 
In this study, we evaluate the quality of generated data from TRecgNet. Since depth data suffers a lot from scarcity and value error problems, we focus on depth data generation. We compare TRecgNet with two image generation approaches. The first one is to generate depth data only using $L1$ loss on pixel-level intensity, which is very common in images reconstruction works~\cite{laina2016deeper} as stated before. We also test images generated by pixel-to-pixel GAN supervision~\cite{isola2017image}. We qualitatively compare some generated examples in Figure~\ref{fig:compare-images}. Images from pixel intensity supervision tend to be blurred stiffly imitating the training data while our TRecgNet produce more natural images, especially for depth data, even for cases that the ground truth is of significant errors, see the bottom row. Interestingly, RGB images generated based on GAN shows an impressive effect on color diversity. We also quantitatively evaluate the quality of generated images by fine tuning pre-trained ResNet18 only using generated images by different methods. Table~\ref{tab:gen_acc} shows the results. Depth images from our TRecgNet outperform other methods by a big margin, indicating its effectiveness for depth data augmentation while GAN based methods achieve a better result on RGB images.

\textbf{Comparison with state-of-the-art methods.}
We report TRecgNet on SUN RGB-D test set compared with state-of-the-art methods, as shown in Table~\ref{tab:sun-rgbd}. Most RGB-D scene recognition methods build upon models pre-trained on Places dataset~\cite{zhou2014learning}. Apart from that, we also report the results using ImageNet pre-trained weights. Most of these methods rely on fine tuning Places-CNN. Song~\etal~\cite{song2018learning} work on learning more effective depth representations from supervised depth patches via SSP~\cite{he2014spatial} and makes comprehensive investigations on fusion strategies, such as multi-scale and aided from object detection. $DF^{2}$Net adopts triplet loss based metric learning~\cite{Li2018DF2NetDF} to learn discriminative and correlative features for modality-specific representation and fusion learning. Our TRecgNets outperform other state-of-the-art methods on both kinds of modalities and their fusion by an evident margin. It is worth noting that our method succeeds in learning modality-specific features from cross-modal transfer learning, and we do not rely much on any sophisticated and specifically designed fusion strategy. 
\subsection{Results on NYUD2 dataset}
We also evaluate TRecgNet on the NYUD2 test set and compare with other representative works. NYUD2 is a relatively small RGB-D dataset, thus we only evaluate the TRecgNet using pre-trained weights of Places dataset or SUN RGB-D dataset. In particular, we study the generalization ability of learned TRecgNet representations on SUN RGB-D. We transfer the learned TRecgNets from SUN RGB-D and fine tune on data from the NYUD2 dataset. We report the results in Table~\ref{tab:nyud2}. We find that for RGB modality, our RGB TRecgNet only yields a slightly better result. We argue that it is mainly because the size of NYUD2 dataset is too small. Negative effects from errors of depth maps badly affects the translation from the RGB to the depth. However, transferring pre-trained weights from SUN RGB-D datasets encourages the TRecgNet to overcome this problem, enhance modality-specific representation power and improve its final recognition performance. When adding the generated data, we observe further promotions for both modalities. Experiments in NYUD2 dataset reveal TRecgNet's requirement on the scale of training data to some degree. Translation using small dataset is difficult in benefiting the multi-training task, especially for modalities that exist measuring errors that can't be ignored. Figure~\ref{fig:layer-acc} also shows some hints for problem that when training TRecgNet in the first several epochs, it tends to behave more unstably than that of fine tuning the backbone network directly and get suboptimal results.

\section{Conclusion and Future Work}
In this paper, we have presented an effective Translate-to-Recognize Network (TRecgNet) to learn modality-specific RGB-D representation for RGB-D scene recognition task. TRecgNet enables a CNN classification network to learn more discriminative feature by a translation process learning essential similarity with cross-modal data. Training TRecgNet allows using unlabeled RGB-D data as initialization which makes up for the data scarcity problem. As experiments demonstrated on SUN RGB-D and NYUD2 datasets, we both achieve the state-of-the-art results validating the effectiveness of the proposed method. In the future, we plan to try instantiate more as well as deeper CNN models such as ResNet50 and VGG Network. We will also try to handle the big error problem of depth data. 
\section*{Acknowledgments} This work is supported by the National Science Foundation of China under Grant No.61321491, and Collaborative Innovation Center of Novel Software Technology and Industrialization.

{\small
\bibliographystyle{ieee_fullname}
}

\end{document}